\documentclass[10pt,journal,final,finalsubmission,twocolumn]{IEEEtran}
\usepackage[sort&compress,numbers]{natbib}
\usepackage{graphicx}
\usepackage{epstopdf}
\usepackage{subfigure}
\usepackage{caption}
\usepackage{algorithm}
\usepackage{algorithmic}
\usepackage{multirow}
\ifCLASSINFOpdf
\else
\fi
\begin{document}
%
% paper title
% can use linebreaks \\ within to get better formatting as desired
% Do not put math or special symbols in the title.
\title{Fingerprint Classification Based on Depth Neural Network}
%
%
% author names and IEEE memberships
% note positions of commas and nonbreaking spaces ( ~ ) LaTeX will not break
% a structure at a ~ so this keeps an author's name from being broken across
% two lines.
% use \thanks{} to gain access to the first footnote area
% a separate \thanks must be used for each paragraph as LaTeX2e's \thanks
% was not built to handle multiple paragraphs
%

\author{Ruxin~Wang,%~\IEEEmembership{Member,~IEEE,}
        ~Congying~Han,
        ~Yanping~Wu,%~\IEEEmembership{Fellow,~OSA,}
        ~and~Tiande~Guo%~\IEEEmembership{Life~Fellow,~IEEE}% <-this % stops a space

%\thanks{This work was funded by the Chinese National Natural Science Foundation (11331012, 71271204, 11101420).}
\thanks{R. Wang, C. Han and T. Guo are with the School
of Mathematical Science, University of Chinese Academy of Sciences, Beijing 100049,
China (e-mail: wangruxin12@mails.ucas.ac.cn; hancy@ucas.ac.cn; tdguo@ucas.ac.cn).}% <-this % stops a space
\thanks{Y. Wu is with the School of Computer and Control Engineering, University of Chinese Academy of Sciences, Beijing 100049,
China (e-mail: wuyanping0711@163.com).}% <-this % stops a space
}
\maketitle

% As a general rule, do not put math, special symbols or citations
% in the abstract or keywords.
\begin{abstract}
Fingerprint classification is an effective technique for reducing the candidate numbers of fingerprints in the stage of matching in automatic fingerprint identification system (AFIS). In recent years, deep learning is an emerging technology which has achieved great success in many fields, such as image processing, natural language processing and so on. In this paper, we only choose the orientation field as the input feature and adopt a new method (stacked sparse autoencoders) based on depth neural network for fingerprint classification. For the four-class problem, we achieve a classification of 93.1 percent using the depth network structure which has three hidden layers (with 1.8\% rejection) in the NIST-DB4 database. And then we propose a novel method using two classification probabilities for fuzzy classification which can effectively enhance the accuracy of classification. By only adjusting the probability threshold, we get the accuracy of classification is 96.1\% (setting threshold is 0.85), 97.2\% (setting threshold is 0.90) and 98.0\% (setting threshold is 0.95). Using the fuzzy method, we obtain higher accuracy than other methods.
\end{abstract}

% Note that keywords are not normally used for peerreview papers.
\begin{IEEEkeywords}
deep learning, fingerprint classification, fuzzy classification, multi-classifier, sparse autoencoder.
\end{IEEEkeywords}

% For peer review papers, you can put extra information on the cover
% page as needed:
% \ifCLASSOPTIONpeerreview
% \begin{center} \bfseries EDICS Category: 3-BBND \end{center}
% \fi
%
% For peerreview papers, this IEEEtran command inserts a page break and
% creates the second title. It will be ignored for other modes.
\IEEEpeerreviewmaketitle

\section{Introduction}
% The very first letter is a 2 line initial drop letter followed
% by the rest of the first word in caps.
%
% form to use if the first word consists of a single letter:
% \IEEEPARstart{A}{demo} file is ....
%
% form to use if you need the single drop letter followed by
% normal text (unknown if ever used by IEEE):
% \IEEEPARstart{A}{}demo file is ....
%
% Some journals put the first two words in caps:
% \IEEEPARstart{T}{his demo} file is ....
%
% Here we have the typical use of a "T" for an initial drop letter
% and "HIS" in caps to complete the first word.
\IEEEPARstart{C}{lassification} of fingerprint images plays an important role in automatic fingerprint identification system (AFIS), especially in the big data times. With the maturing of fingerprint identification technology and the expanding of fingerprint database, the accuracy and speed of fingerprint identification are required higher and higher. Fingerprint classification is a method which can effectively reduce the size of the database. According to Henry (1900) \cite{bg1}, the fingerprint is divided into five categories: the left loop, right loop, arch, tented arch and the whorl (Fig.1). So, in the stage of query, the query fingerprint compares with the other fingerprints which have the same class directly.
% You must have at least 2 lines in the paragraph with the drop letter
% (should never be an issue)

\begin{figure*}[t]
\centering \subfigure[]{\label{fig:a}\includegraphics[width=0.28\textwidth]{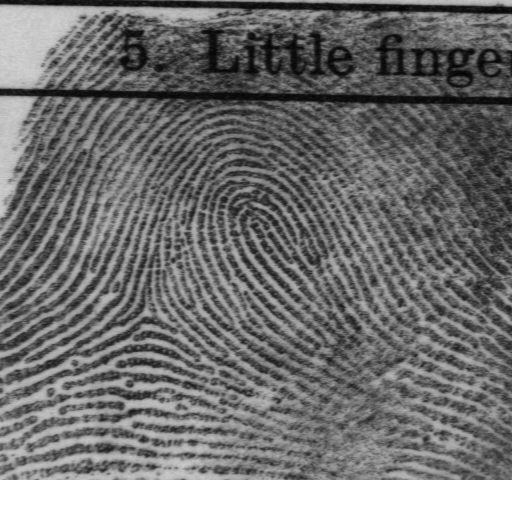}}
\hfill \subfigure[]{\label{fig:b}\includegraphics[width=0.28\textwidth]{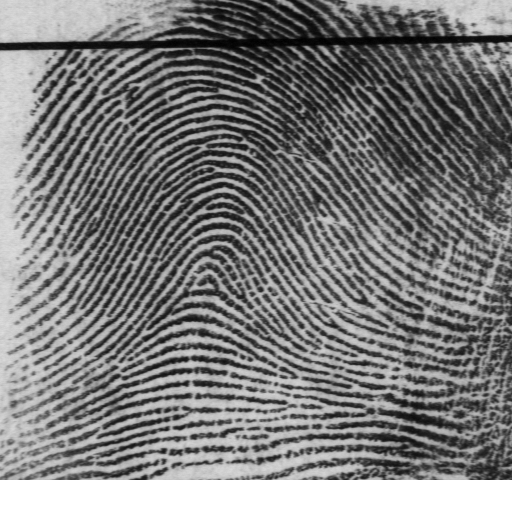}}
\hfill \subfigure[]{\label{fig:c}\includegraphics[width=0.28\textwidth]{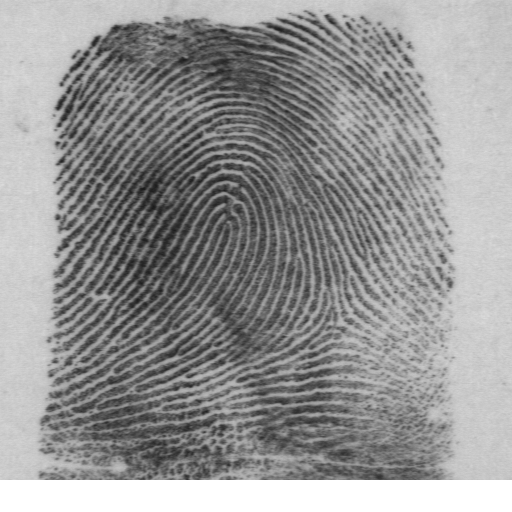}}
\hfill \subfigure[]{\label{fig:d}\includegraphics[width=0.28\textwidth]{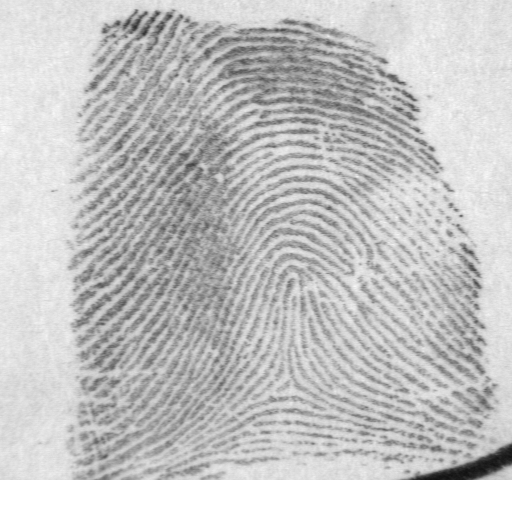}}
\hfill \subfigure[]{\label{fig:e}\includegraphics[width=0.28\textwidth]{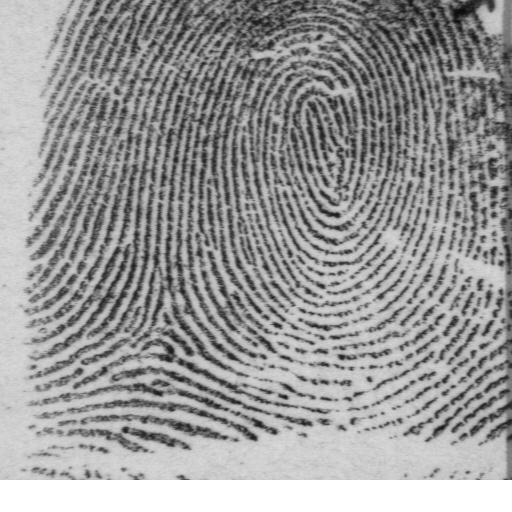}}
\hfill \subfigure[]{\label{fig:f}\includegraphics[width=0.28\textwidth]{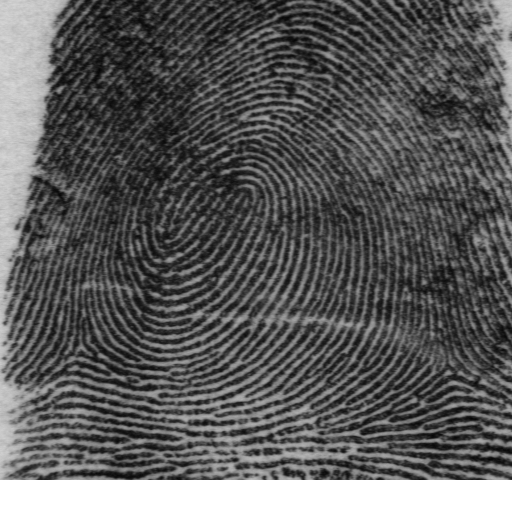}}
\caption{Six samples from NIST-DB4 database. (a) right loop, (b) arch, (c) left loop, (d)
tented arch, (e) whorl, (f) twin loop (labeled as whorl in NIST-DB4 database).
} \label{fig:1}
\end{figure*}

In the past decades, people have put forward a lot of fingerprint classification algorithms, which can be divided into four types \cite{bg2}: (a) rule-based methods, this kind of algorithms usually simply classify the fingerprint according to the number and position of the singularities. The usual practice is that the singular points are detected firstly, and then a coarse decision is made according to the position and the type of them. For example, in Kawagoe et al. (1984) \cite{bg3}, they adopt this technology. (b) syntactic approaches, a classical algorithm is presented by Rao et al \cite{bg4}, which they analyze the pattern of ridge line flow according to the directional image and then classify the fingerprint based on some grammars. (c) structural approaches, structural approaches usually use relational trees or relational graphs for integrating ¡°features¡±, for instance, in Maio and Maltoni (1996) \cite{bg5}, ¡°homogeneous¡± orientation regions are selected as the characteristics, then a relational graph model is adopted for describing the feature structure. (d) traditional neural network-based methods, for example, Bowen et al. (1992) \cite{bg6} propose a pyramidal architecture for fingerprint classification. In their methods, the location of the singularities and the corresponding local direction image are used for training the networks, then the outputs through another network \cite{bg7}, which produces the final classification. Although these methods have achieved some good results, singular points information and the fingerprint alignment are usually used by many methods \cite{bg8} - \cite{bg10}. So the number and the position of singular points will directly affect the final classification results, a good algorithm for detecting singular points is the key to this kind of methods for fingerprint classification. The the error information of singular points often leads to the failure of classification. We have proposed some good algorithms for detecting singular points, for example Fan and Guo et al. (2008) \cite{bg11} combine the Zero-pole Model and Hough Transform to detect singular points, which performs well and makes the result more robust to noise.

It is known that the theory and application of deep learning have achieved great success in recent years. All of the machine-learning algorithms need to have the characteristics as the input, a good feature representation can be thought as a key to an algorithm. The traditional definition and selection of feature are often completed by artificial. But the manual features selection is often time-consuming, and the feature usually has certain subjectivity, but also requires some prior knowledge. However, by constructing a depth neural network which imitates the cognitive process from observation to abstraction by human brain, some disadvantages of traditional artificial features will be overcome.  Through this kind of layered structure, deep learning achieves a very good result in feature self-taught learning and obtains ¡°multi-scale¡± characteristics. Hinton (2006) \cite{bg12} uses the restricted Boltzmann machines (RBMS) to learn low-dimensional codes that work much better than PCA as a tool to reduce the dimensionality of data. Yann LeCun et al (1998) \cite{bg13} apply the convolutional neural network to the handwriting recognition, which acquires a very good result. Similarly, Pascal Vincent et al. (2008) \cite{bg14} propose the stacked denoising autoencoders model and also show the surprising advantage in handwritten recognition experiment. Although there are a variety of deep structure models \cite{bg15}, they all gain very good results in many fields, such as visual \cite{bg16} - \cite{bg18}, natural language processing \cite{bg19}, \cite{bg20} and signal processing and so on. So we believe that the method based on depth neural network can also achieve good results for the fingerprint classification.

In this paper, we  adopt a deep network structure, the sparse autoencoder neural network (containing three hidden layer) \cite{bg21}, to ¡°learn¡± the "pattern characteristic" of each class of the fingerprint by unsupervised feature self-taught learning, then the features obtained by sparse autoencoder via a multi classifier for fingerprint classification. In order to further improve the accuracy of the classification, we propose the fuzzy classification method. Section \uppercase\expandafter{\romannumeral2} gives a brief introduction about the sparse autoencoder. And in section \uppercase\expandafter{\romannumeral3}, we introduce the softmax regression. Section \uppercase\expandafter{\romannumeral4} a fuzzy method for fingerprint classification is described. Section \uppercase\expandafter{\romannumeral5} presents experimental results for classification and fuzzy classification. Section \uppercase\expandafter{\romannumeral6} gives the conclusion for the paper.

\section{Unsupervised Feature Self-Taught Learning}
In this section, we introduce the unsupervised feature self-taught learning algorithm---the sparse autoencoder firstly and then we propose our features which are used for training the autoencoder. We demonstrate that our features are more effective than others by the theory and experiment. The results of training are satisfied.

\subsection{Unsupervised Feature Self-Taught Learning Model}
A learning algorithm based on deep structure, namely the deep learning theory, has been proved that it achieves very beautiful result in many fields, especially in computer vision, image processing, audio processing and natural language processing. At the same time, the related theory of deep learning has also been developing and updating. As a method of unsupervised feature self-taught learning, this kind of hierarchical structure can achieve the feature dimension reduction and self-selection effectively \cite{bg22}. Hinton et al. (2006) \cite{bg11} use the RBMs for dimension compression and gain a very good result. In this section, we will introduce another self-taught learning method---the sparse autoencoder model. The sparse autoencoder learning algorithm is one approach to automatically learn features from unlabeled data. It is a variant of basic autoencoder \cite{bg23}, \cite{bg24}. Unlike the traditional neural network, an autoencoder neural network is an unsupervised learning algorithm that applies back propagation, setting the target values to be equal to the inputs in network terminal. Given an unlabeled training examples $\{x^{(1)},...,x^{(m)}\}$,The basic autoencoder can be written as follows:

\begin{equation}	
J_{ae}=\arg\min_{W,b}\frac{1}{m}\sum_i||h_{W,b}(x^{(i)})-x^{(i)}||^2_2
\end{equation}

\begin{figure*}[htbp]
\centering
\subfigure{\includegraphics[width=4.5in]{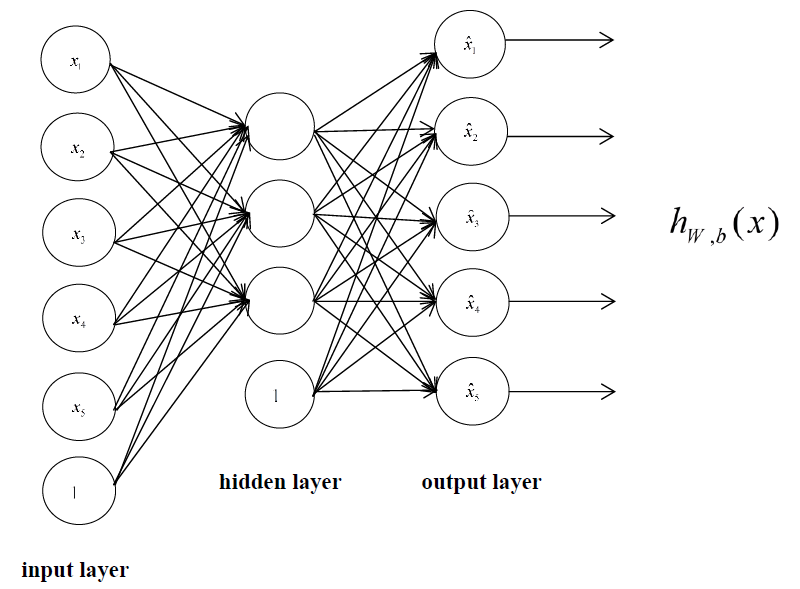}}
\caption{A sparse autoencoder with five input nodes, three hidden nodes and five output nodes.}
\label{fig_sim}
\end{figure*}

In order to explore more "effective" composition in the input data, we impose the sparsity constraint on the autoencoder network, so that it can obtain sparse autoencoder neural network which we mentioned at the beginning. Take one hidden layer sparse autoencoder network for example (Fig.2), the objective function is as follows:

\begin{eqnarray}
J_{sae}(W,b)=\arg\min_{W,b}\frac{1}{m}\sum_i||h_{W,b}(x^{(i)})-x^{(i)}||^2_2\nonumber\\+\lambda\sum_l\sum_{i,j}(W^{l}_{i,j})^2+\beta\sum_iKL(\rho||\hat\rho_j)
\end{eqnarray}

\begin{equation}
KL(\rho||\hat\rho_j)=\rho\log\frac{\rho}{\hat\rho_j}+(1-\rho\log\frac{1-\rho}{1-\hat\rho_j})
\end{equation}

Where in the equation, the first term is the basic autoencoder and the second term is the regularization term, which helps prevent over-fitting. $(W,b)$ are the parameters of the neural network. $\lambda,\beta>0$ are the balance parameters. $KL(\rho||\hat\rho_j)$  is the Kullback-Leibler $(KL)$ divergence between a Bernoulli random variable with mean $\rho$  and a Bernoulli random variable with mean $\hat\rho_j$, $\rho$  is a sparsity parameter, typically a small value close to zero, $\hat\rho_j=\frac{1}{m}\sum_i[a_j(x^{(i)})]$ is the average activation of hidden unit $j$. $a_j(x^{(i)})$ denotes the activation of this hidden unit $j$.

%\begin{figure}[!t]
%\centering
%\includegraphics[width=2.5in]{myfigure}
% where an .eps filename suffix will be assumed under latex,
% and a .pdf suffix will be assumed for pdflatex; or what has been declared
% via \DeclareGraphicsExtensions.
%\caption{Simulation Results.}
%\label{fig_sim}
%\end{figure}

\begin{figure*}[!t]
\centering
\subfigure{\includegraphics[width=4.5in]{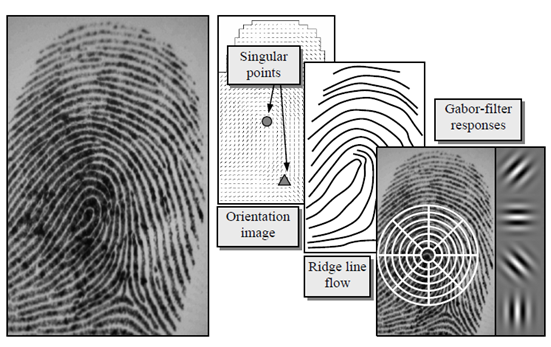}}
\caption{Some classification feature examples from the literature \cite{bg2}.}
\label{fig_sim}
\end{figure*}
So, we obtain a sparse autoencoder model. It can effectively realize feature extraction and dimension reduction of the input data. The number of the intermediate hidden layers can be set based on the actual needs. That way with the increase number of the hidden layers, the deep structure becomes more and more obvious.

% needed in second column of first page if using \IEEEpubid
%\IEEEpubidadjcol
\subsection{Feature Selection}
\subsubsection{Selecting Feature For Classification}
After choosing the suitable learning algorithm, the next step is the unsupervised training. So the feature selection becomes the second problem we will face to. For fingerprint classification, there are a lot of researches on it in that the difficulty and importance of its own. Although a large number of classification algorithms have been proposed, a relatively small number of features extracted from fingerprint images have been used \cite{bg2}. In particular, almost all the methods are based on one or more of the following features: ridge line flow, orientation image, singular points, and Gabor filter responses (Fig.3).

Here, we choose the orientation field as our feature. There are some reasons: the first one is that the orientation field belongs to global feature of a fingerprint, which can effectively express the ridge flow pattern and the ridge type. If computed with sufficient accuracy and detail, the orientation field contains all the information required for the classification. Secondly, due to the effect of detecting the number and the location of singular points, the classification result is quite sensitive, such as wrong in detecting the number of singular points would lead to misclassification, what's more, it also influences the processing speed. Thirdly, the Gabor filter responses are usually employed as a coding technique of ¡°center point¡± by FingerCode-based method \cite{bg25} - \cite{bg27}. In conclusion, we only choose the orientation information as the input feature for learning.

\subsubsection{Computing The Orientation Field}
Computing the orientation field is a crucial step for constructing the AFIS. The accurate orientation field has a very important effect on extracting the minutiae points and fingerprint matching. Also, for our method, it is very important for using a good algorithm for computing it, because we choose the orientation field as the only input feature.

The classical method for obtaining the orientation field is the gradient-based method, for example, Rao et al. (1990) \cite{bg28} use the block direction instead of every point direction inside the block. Relative to the point direction, the block direction has high ability of anti-noise. In recent years, some new methods spring up, for instance, Feng and Jain et al. (2013) \cite{bg29} propose a method based on dictionary learning for estimating the orientation field of latent fingerprints. Also we have proposed some methods for orientation field estimation, for example, Wu and Guo et al. (2013) \cite{bg30} propose a SVM-based method for fingerprint and palmprint orientation field estimation. Shao and Han et al. (2012) \cite{bg31} use the nonnegative matrix factorization (NMF) to initialize the fingerprint orientation field instead of the gradient-based approach, which obtains more robust results.

%a) Divide $G$ into blocks of size $w\times w$, we choose $w=20$ in our algorithm.
%
%b) Compute the gradients $G_x(i,j)$ and $G_y(i,j)$ at each pixel $(i,j)$.The gradient operator can choose the Sobel operator.
%
%c) Estimate the local orientation of each block centered at pixel   using the following equations:
%
%\begin{equation}
%[G_{Bx},G_{By}]_{(i,j)}^T=[\sum_{i=1}^w\sum_{j=1}^wG_{Sx}(i,j),\sum_{i=1}^w\sum_{j=1}^wG_{Sy}(i,j)]^T
%\end{equation}
%
%\begin{eqnarray}
%\theta=\frac{1}{2}\pi+\frac{1}{2}\left\{
%\begin{array}{rcl}
%\arctan\frac{G_{By}}{G_{Bx}},     {G_{Bx}\geq 0\bigcap G_{By} unrestraint}\\
%\\
%\arctan\frac{G_{By}}{G_{Bx}}+\pi,      {G_{Bx}\leq 0\bigcap G_{By}\geq 0}\\
%\\
%\arctan\frac{G_{By}}{G_{Bx}}-\pi,      {G_{Bx}\leq 0\bigcap G_{By}\leq 0}\\
%\end{array} \right.
%\end{eqnarray}
%
%where \begin{eqnarray}\left(
%  \begin{array}{ccc}
%    G_{Sx}(i,j) \\
%    G_{Sy}(i,j) \\
%  \end{array}
%\right)=\left(
%  \begin{array}{ccc}
%    G_x(i,j)^2-G_y(i,j)^2\\
%    2G_x(i,j)^2G_y(i,j)^2 \\
%  \end{array}
%\right).\end{eqnarray}

In our algorithm, we use our orientation field estimation method for extracting the orientational feature, which is currently in use in our AFIS. However, we do not use the each block direction as the input feature directly, instead, we use the trigonometric function for converting the direction value $\theta$ into $(\sin 2\theta,\cos 2\theta)^T$. Without loss of generality, we compare these features through an experiment. We choose 2000 samples as the training set and 400 samples as the test set from the NIST-DB4 database for testing the influences of different input features for fingerprint classification. The direction angle $\theta$, the sine and cosine value of direction angle and the any two combined vector of $\sin 2\theta$, $\cos 2\theta$ and $\theta$ are adopted in our test experiment and we use three kinds of network structure with different hidden layers for testing. The results show that using combined vector $(\sin 2\theta,\cos 2\theta)^T$ as the input feature can achieve better results for classification than other features (Fig.4).

Through the results, we find that the combined features have higher accuracy than single feature. For the single input feature, the cosine value of direction angle as the input feature has the worst performance for classification, and the sine value of direction angle and the direction angle have no obvious difference for classification. But for the combined features, the features combined $\cos 2\theta$ with $\theta$ have better performance than the features combined $\sin 2\theta$ with $\theta$. We can give a unified explanation from the perspective of information theory for the above experimental phenomena. We use the information gain \cite{bg32} (using Algorithm 1) for describing the influence of the features for classification.

\begin{algorithm}%
[htb]
\caption{}
\label{alg:1}
\begin{algorithmic}[1]
\STATE Input training set $T$ and features $F$
\STATE Compute the empirical entropy $H(T)$ of $T$:

$H(T)=-\sum_{k=1}^K\frac{|C_k|}{|T|}\log\frac{|C_k|}{|T|}$
\STATE Compute the empirical conditional entropy $H(T|F)$:

$H(T|F)=\sum_{i=1}^n\frac{|T_i|}{|T_i|}$
\STATE Compute the information gain $g(T,F)$:

$g(T,F)=H(T)-H(T|F)$

\STATE Output the information gain $g(T,F)$, where $|.|$ indicates sample capacity, $C_k$ indicates $k$ categories.
\end{algorithmic}
\end{algorithm}

According to information theory, the feature with high information gain has stronger power for classification. We can gain that $(\sin 2\theta,\cos 2\theta)^T$ has greater information gain than others and it can eliminate the direction errors due to the circularity of angles. At the same time, we can obtain that the information gain of $(\cos 2\theta,\theta)^T$ is higher than $(\sin 2\theta,\theta)$.

So, from both theory and experiment (Fig.4), we choose the $(\sin 2\theta,\cos 2\theta)^T$ as the input features for training the neural network.

\begin{figure*}[t]
\centering \subfigure[]{\label{fig:a}\includegraphics[width=0.3\textwidth]{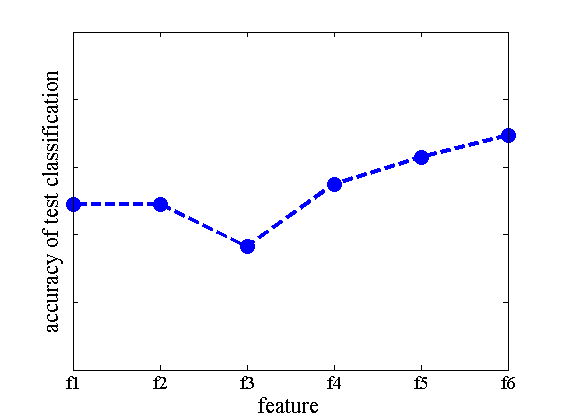}}
\hfill \subfigure[]{\label{fig:b}\includegraphics[width=0.3\textwidth]{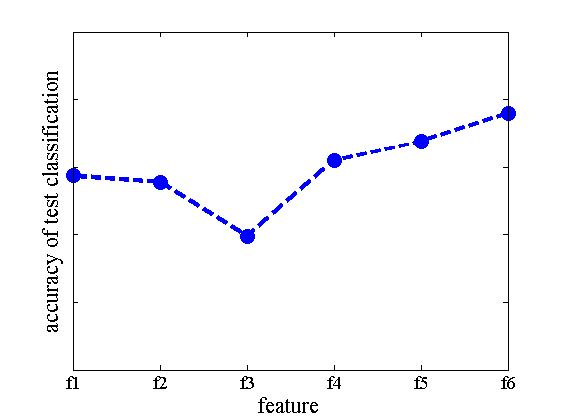}}
\hfill \subfigure[]{\label{fig:c}\includegraphics[width=0.3\textwidth]{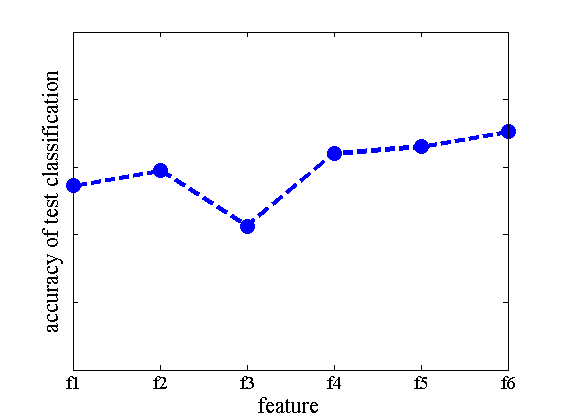}}
\caption{The results of feature selecting experiment in (a) one hidden layer, (b) two hidden layers and (c) three hidden layers network. Where f1 := $\theta$, f2 := $\sin2\theta$, f3 := $\cos2\theta$, f4 := $(\sin2\theta,\theta)^T$, f5 := $(\cos2\theta,\theta)^T$, f6 := $(\sin2\theta,\cos2\theta)^T$.
} \label{fig:4}
\end{figure*}

\begin{figure*}[htbp]
\centering \subfigure[]{\includegraphics[width=0.34\textwidth]{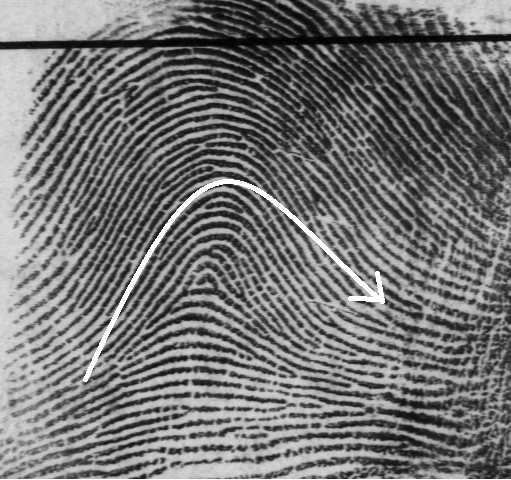}}
\hfill \subfigure[]{\includegraphics[width=0.32\textwidth]{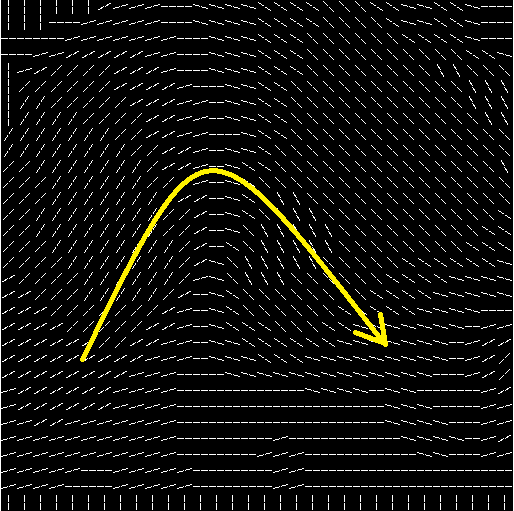}}
\hfill \subfigure[]{\includegraphics[width=0.32\textwidth]{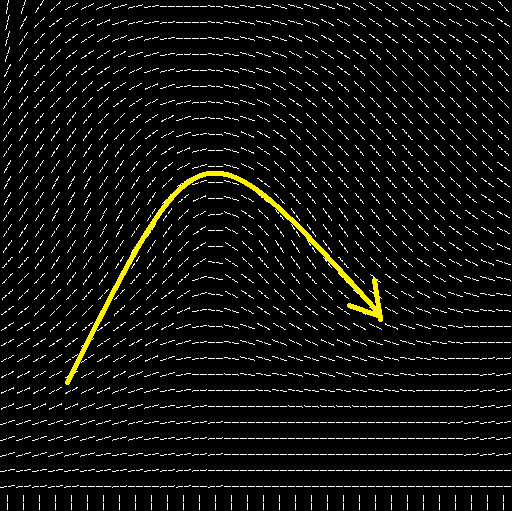}}
\caption{The study results: (a) arch, (b) is the input orientation field, (c) is the reconstructed orientation field by sparse autoencoder.} \label{fig:5}
\end{figure*}

\begin{figure*}[htbp]
\centering \subfigure[]{\includegraphics[width=0.34\textwidth]{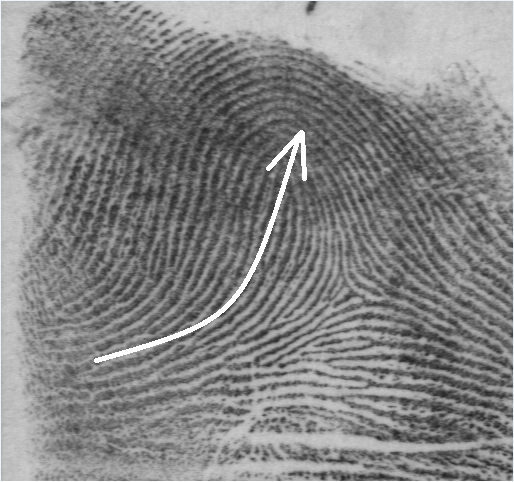}}
\hfill \subfigure[]{\includegraphics[width=0.32\textwidth]{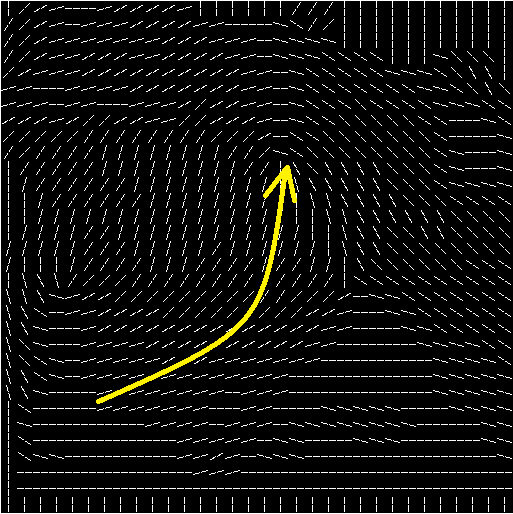}}
\hfill \subfigure[]{\includegraphics[width=0.32\textwidth]{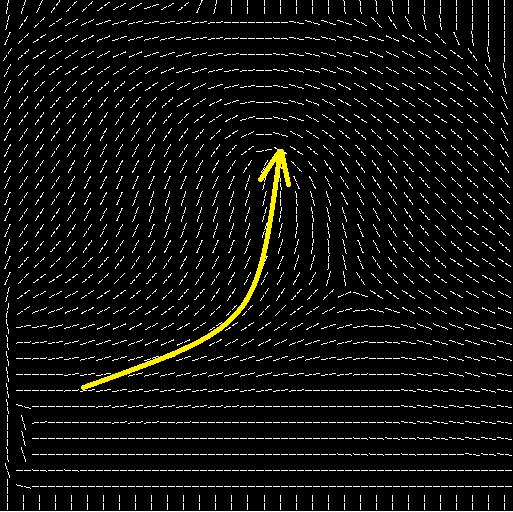}}
\caption{The study results: (a) left loop, (b) is the input orientation field, (c) is the reconstructed orientation field by sparse autoencoder.} \label{fig:5}
\end{figure*}

\begin{figure*}[htbp]
\centering \subfigure[]{\includegraphics[width=0.34\textwidth]{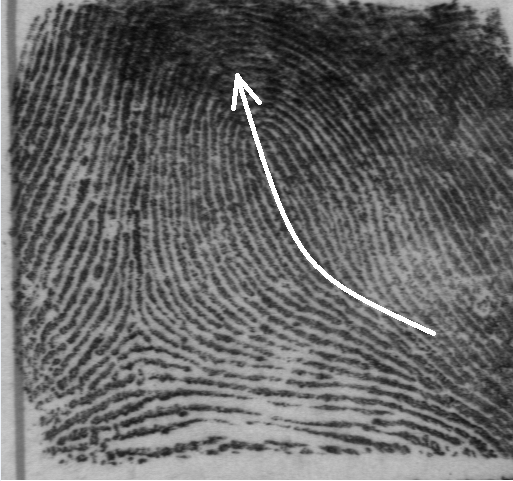}}
\hfill \subfigure[]{\includegraphics[width=0.32\textwidth]{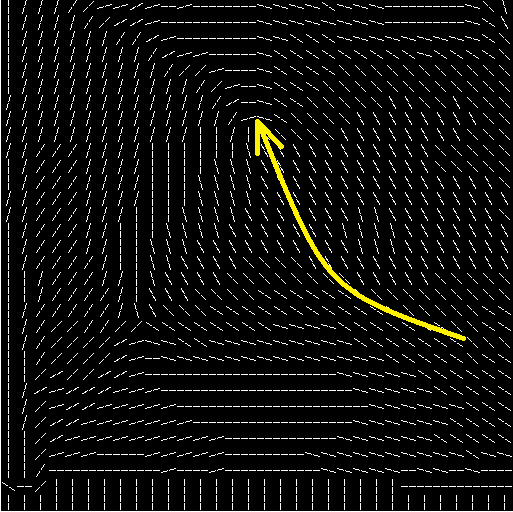}}
\hfill \subfigure[]{\includegraphics[width=0.32\textwidth]{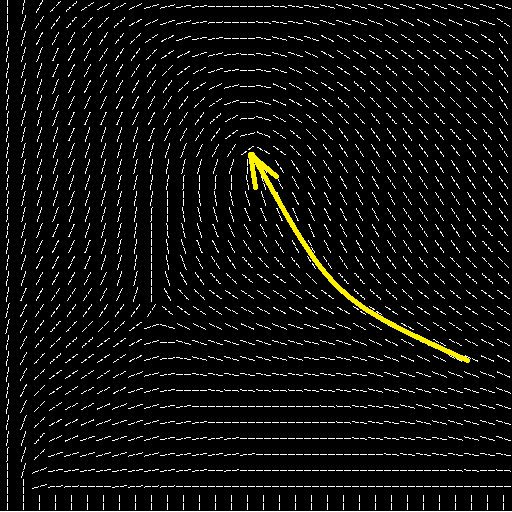}}
\caption{The study results: (a) right loop, (b) is the input orientation field, (c) is the reconstructed orientation field by sparse autoencoder.} \label{fig:5}
\end{figure*}

\begin{figure*}[htbp]
\centering \subfigure[]{\includegraphics[width=0.34\textwidth]{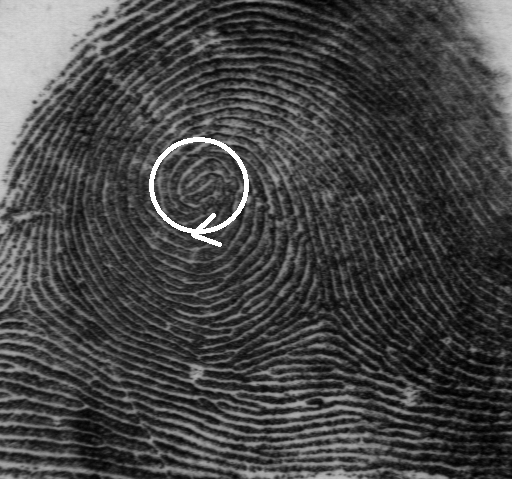}}
\hfill \subfigure[]{\includegraphics[width=0.32\textwidth]{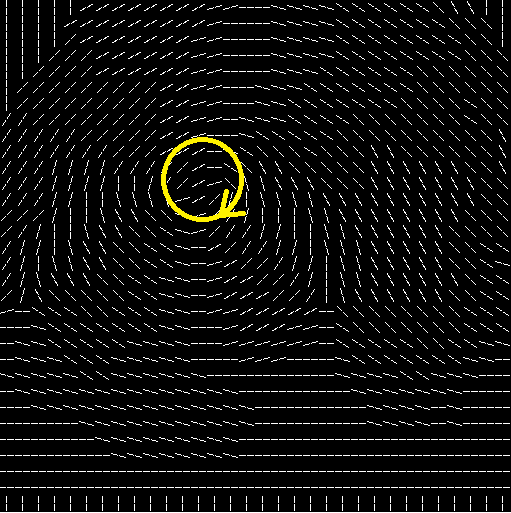}}
\hfill \subfigure[]{\includegraphics[width=0.32\textwidth]{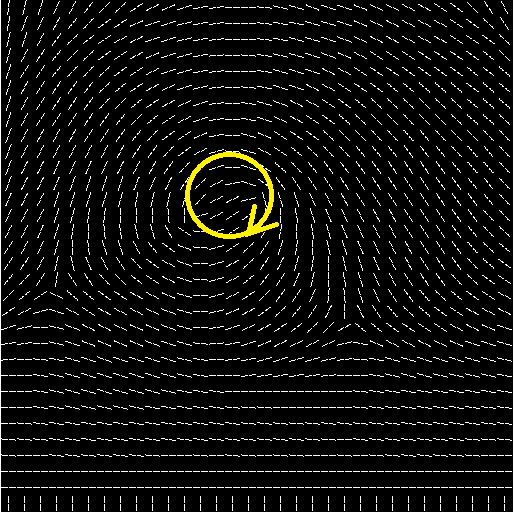}}
\caption{The study results: (a) whorl, (b) is the input orientation field, (c) is the reconstructed orientation field by sparse autoencoder.} \label{fig:5}
\end{figure*}

\begin{figure*}[htbp]
\centering \subfigure[]{\includegraphics[width=0.34\textwidth]{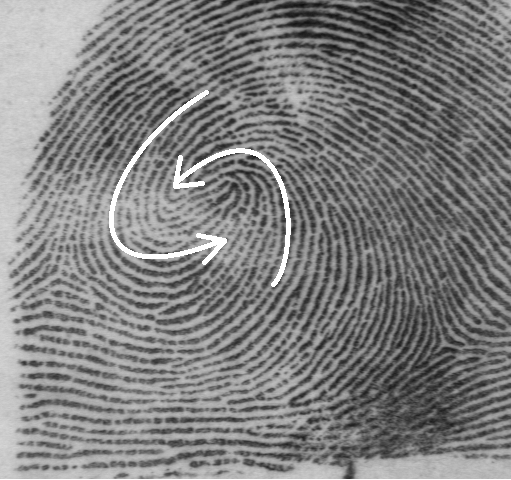}}
\hfill \subfigure[]{\includegraphics[width=0.32\textwidth]{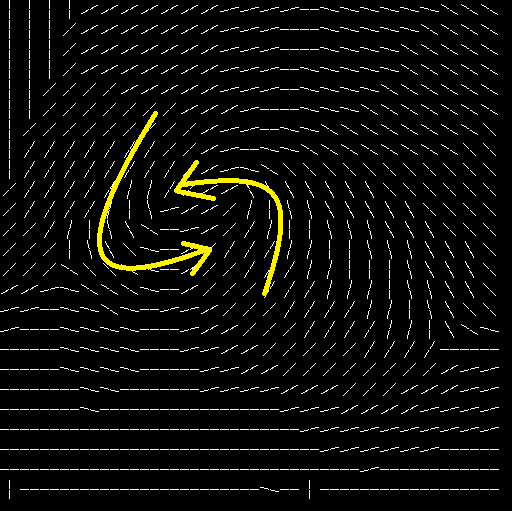}}
\hfill \subfigure[]{\includegraphics[width=0.32\textwidth]{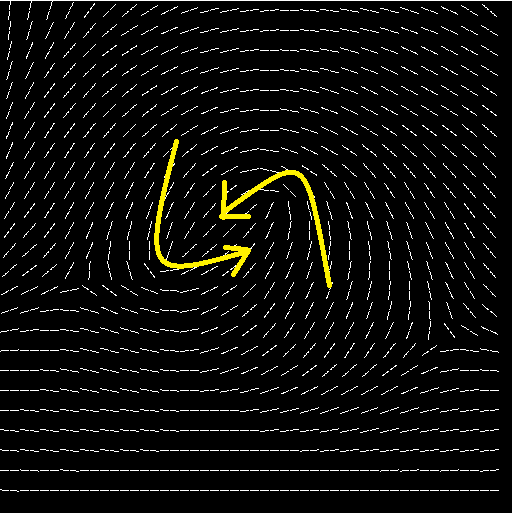}}
\caption{The study results: (a) twin loop whorl, (b) is the input orientation field, (c) is the reconstructed orientation field by sparse autoencoder.} \label{fig:5}
\end{figure*}

\begin{figure*}[!t]
\centering
\subfigure{\includegraphics[width=5in]{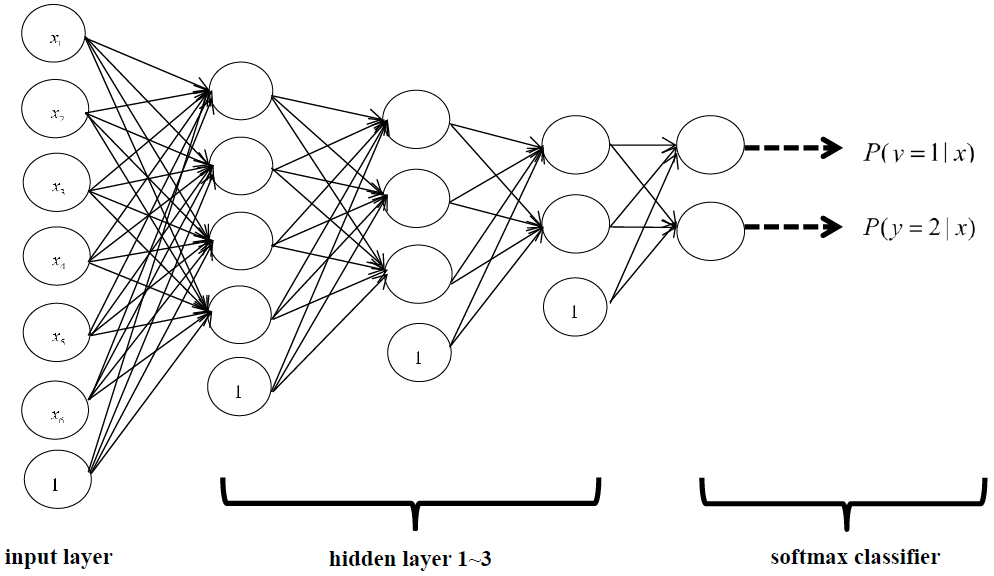}}
\caption{A sketch map of our classification method.}
\label{fig_sim}
\end{figure*}

\subsubsection{Showing The Study Results}
According to the hypothesis of the sparse autoencoder which trys to learn an approximation to the identity function, so as to output $h_{W,b}(x^{(i)})$ that is similar to $x^{(i)}$, we can obtain the features that can reconstruct the input orientation field. The following is a list of the results (Fig.5 - Fig.9):

We hope that the features after training can reflect the types of fingerprints. For this purpose, we need not require too much in computing the orientation field. The results show that the unsupervised learning algorithm and the input feature which we choose are feasible and the loss for expressing types of fingerprint between input and output is no obvious to the naked eye.

\section{Multi-Classifier}
Fingerprint classification is a multi-class classification problem, there are many lectures disposing this problem by designing neural networks¡¯ output directly. Different from the above approach is that here we select the deep neural network to extract each kind of fingerprint "characteristic", and then these features through a multi classifier are used for classification.

Input the fingerprint orientation field and the low dimensional characteristics are stripped out after the unsupervised feature learning. These characteristics are thought to be useful and reliable for the next classifier training in that the characteristics of the sparse autoencoder. We choose the softmax regression model \cite{bg33}, \cite{bg34} a supervised machine learning algorithm, as our multi-classifier.

Softmax regression model generalizes logistic regression to classification problems. It obtains a good result in handwriting recognition problem, where the goal is to distinguish between 10 different numerical digits. Softmax regression model is also regarded as a generalized linear model which is generated from a multinomial distribution. Consider a classification problem in which the class label $y$ can take on any one of $k$ values, so $y\in\{1,...,k\}$. Given a test input $x$, we estimate the probability $p(y=j|x)$ that for each class $j$. Therefore, the model can be express as following:

\begin{eqnarray}
h_\theta(x)=\left[
  \begin{array}{ccc}
    p(y=1|x;\theta) \\
    p(y=2|x;\theta) \\
    \vdots\\
    p(y=k|x;\theta) \\
  \end{array}
\right]=\frac{1}{\sum_{j=1}^ke^{\theta_k^Tx}}\left[
  \begin{array}{ccc}
    e^{\theta_1^Tx}\\
    e^{\theta_2^Tx} \\
    \vdots\\
    e^{\theta_k^Tx} \\
  \end{array}
\right].
\end{eqnarray}

where $\theta_1,\theta_2,...,\theta_k\in R^{n+1}$ are the parameters of our model. Now, given a training set $\{(x^{(1)},y^{(1)}),(x^{(2)},y^{(2)}),...,(x^{(m)},y^{(m)})\}$ of $m$ labeled examples, the model parameters $\theta$ are trained to minimize the cost function:

\begin{eqnarray}
J(\theta)=-\frac{1}{m}\sum_{i=1}^m\sum_{j=1}^kI\{y^{(i)}=j\}\log\frac{e^{\theta_j^Tx^{(i)}}}{\sum_{l=1}^ke^{\theta_l^Tx^{(i)}}}\nonumber\\+\frac{\lambda}{2}\sum_{i=1}^k\sum_{j=0}^n\theta_{ij}^2
\end{eqnarray}

Where $I(.)$ is the indicator function, so that $I(true)=1$, and $I(false)=0$. The second term is a regularization term, $\lambda>0$. There are two advantages introducing the regularization term. The first one is that the regularization term can decrease the magnitude of the parameters, and help prevent over-fitting. The second one is taking care of the numerical problems associated with softmax regression's over-parameterized representation. By adding the regularization term, the cost function is now strictly convex, which can ensure the algorithm can converge to the unique solution, also the global optimal solution. That is also an important reason why we choose the model for the multi-classifier.

We input the features which obtained from the sparse autoencoder to the multi-classifier, and then using the trained classifier can classify the fingerprints (Fig.10).

\begin{figure*}[t]
\centering
\subfigure{\includegraphics[width=5in]{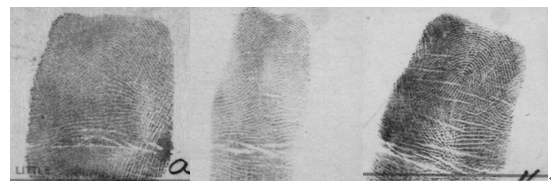}}
\caption{ Three examples of fingerprints with poor quality.}
\label{fig_sim}
\end{figure*}

\begin{figure*}[!t]
\centering
\subfigure{\includegraphics[width=3.5in]{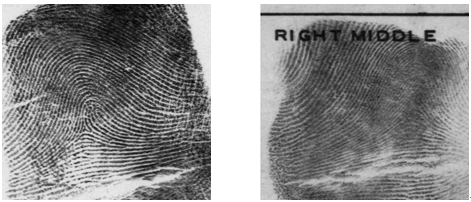}}
\caption{ Two examples of ¡°ambiguous¡± fingerprints.}
\label{fig_sim}
\end{figure*}

\section{Fuzzy Classification}
Through the previous introduction, now a new test fingerprint can be classified by the multi-classifier. In practice, however, people always want to improve the accuracy of classification as far as possible without rising more costs. Fuzzy classification is an effective way to improve the classification accuracy. We are based on the following considerations:

Firstly, any classification system can not guarantee the accuracy rate of 100 percent for the error classifying samples, especially for the images of poor quality and images containing additional lines or tabulations \cite{bg8} (Fig.11). Naturally, these fingerprints would be wrong to the other categories. So, for this part of the loss of accuracy, we can save by fuzzy approach. As long as the fuzzy methods would not bring huge amount of computation, the method is acceptable, but also feasible.

Secondly, some fingerprint is ¡°ambiguous¡± (Fig.12), there are many reasons to cause this kind of "ambiguity", such as apparent injuries or fingerprint ridge structure itself having characteristics of two different types and so on, which could easily cause a wrong classification for an automatic classification system, even for human experts. NIST-DB4 contains 350 ¡°ambiguous¡± fingerprint pairs (about 17\%) \cite{bg35} which are marked by experts. Therefore, in order to avoid potential problems, we divide them into each of the ¡°ambiguous¡± classes for this kind of fingerprints by fuzzy approach.

Based on the above two points, we also adopt softmax regression model for fuzzy classification.

The advantage of softmax regression model is its output, which is represented in probability form, namely, the fingerprint belongs to a class of maximum likelihood. But usually the category which its second probability belongs to is "trusted". For these fingerprints, their output results often have the following characteristics: firstly, the first probability value is small (e.g. less than 0.6), then the gap between second probability value and the first one is very small. Because of the advantages of softmax regression, we use the second probability value of each fingerprint for the fuzzy classification.

\section{Experimental Results}
The fingerprint classification algorithm we proposed was tested on the NIST-DB4 database. The target class is four classes: arch (A), left loop (L), right loop (R) and whorl (W). We combine the arch and tented arch into one category for two main reasons. Firstly, according to statistics, more than 90\% of the fingerprints belong to three classes, namely, left loop, right loop and whorl. And the arch-type fingerprint proportion is very small. So, from a statistical point of view, fingerprints are unevenly distributed among them. Secondly, to define these two classes is difficult, which brings difficulties to the separation. And further more, some fingerprints are ambiguous, which can not be classified even by experts (Fig.12). Therefore, we classify fingerprints into four categories.

\subsection{Database}
NIST-DB4 database is used for testing classification accuracy by most of algorithms, which consist of 4000 fingerprint images (image size is $512\times512$) and the fingerprints have been manually labeled (A/L/R/TA/W). In our algorithm, we choose half of them as the training set which is used for training the parameters of the sparse autoencoder and softmax regression model, the other half as the testing set.
\subsection{Experimental Design}
Firstly, we divide the fingerprints into $20\times20$ blocks. Then the method mentioned in section \uppercase\expandafter{\romannumeral2} was used for computing the direction of each block, which obtains $25\times25$ direction values, and then convert them to the sine and cosine of the double angle which were combined together as the feature of a fingerprint. The dimension of input feature is 1250.

We first test the sparse autoencoder with a single hidden layer containing 600 (400, 200 as control group) nodes for unsupervised training. The result of classification is 90.35\% (90.2\%,89.7\%). The confusion matrix is as following (Table \uppercase\expandafter{\romannumeral1} - \uppercase\expandafter{\romannumeral3}):

\begin{table}[htbp]
\centering
\caption{\label{comparison}The result using one hidden layer with 200 nodes}
\begin{tabular}{c|c|c|c|c}
\hline
\hline
\multirow{2}{*}{True class 89.7\%} & \multicolumn{4}{|c}{Assigned class}  \\
\cline{2-5}
 & A & L & R & W  \\
\hline
\hline
 A & 768 & 19 & 12 & 1  \\
 \hline
 L & 63 & 321 & 0 & 16  \\
 \hline
 R & 55 & 0 & 335 & 10  \\
 \hline
 W & 7 & 10 & 13 & 370  \\
\hline
\end{tabular}
\end{table}

\begin{table}[htbp]
\centering
\caption{\label{comparison}The result using one hidden layer with 400 nodes}
\begin{tabular}{c|c|c|c|c}
\hline
\hline
\multirow{2}{*}{True class 90.2\%} & \multicolumn{4}{|c}{Assigned class}  \\
\cline{2-5}
 & A & L & R & W  \\
\hline
\hline
 A & 751 & 28 & 18 & 3  \\
 \hline
 L & 52 & 338 & 0 & 10  \\
 \hline
 R & 42 & 1 & 345 & 12  \\
 \hline
 W & 5 & 13 & 12 & 370  \\
\hline
\end{tabular}
\end{table}

\begin{table}[tbp]
\centering
\caption{\label{comparison}The result using one hidden layer with 600 nodes}
\begin{tabular}{c|c|c|c|c}
\hline
\hline
\multirow{2}{*}{True class 90.35\%} & \multicolumn{4}{|c}{Assigned class}  \\
\cline{2-5}
 & A & L & R & W  \\
\hline
\hline
 A & 767 & 21 & 11 & 1  \\
 \hline
 L & 52 & 336 & 1 & 11  \\
 \hline
 R & 57 & 0 & 335 & 7  \\
 \hline
 W & 6 & 9 & 16 & 369  \\
\hline
\end{tabular}
\end{table}

We adjust the number of nodes in hidden layer and obtain that the changes of accuracy get smaller with increasing the number of nodes keeping other parameters of model unchanged (Fig.14). Considering the about $\pm0.2\%$ error of classification since the numerical calculation, we set the first hidden layer with 400 nodes in the next experiments.

\begin{figure}[htbp]
\centering
\subfigure{\includegraphics[width=3.5in]{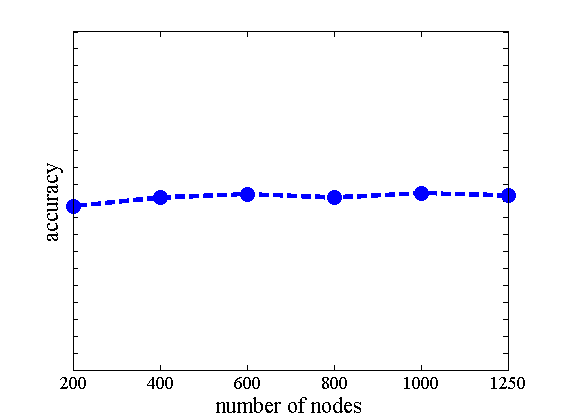}}
\caption{ Relation graph of number of nodes and accuracy in autoencoder with one hidden layer.}
\label{fig_sim}
\end{figure}

Then we add the number of hidden layers (two hidden layers with the nodes 400-100 in our system), the accuracy of classification is improved to about 90.9\%. The confusion matrix is in Table \uppercase\expandafter{\romannumeral4}:

\begin{table}[htbp]
\centering
\caption{\label{comparison}The result using two hidden layers with 400-100 nodes}
\begin{tabular}{c|c|c|c|c}
\hline
\hline
\multirow{2}{*}{True class 90.9\%} & \multicolumn{4}{|c}{Assigned class}  \\
\cline{2-5}
 & A & L & R & W  \\
\hline
\hline
 A & 767 & 17 & 16 & 0  \\
 \hline
 L & 49 & 336 & 2 & 13  \\
 \hline
 R & 54 & 1 & 338 & 7  \\
 \hline
 W & 7 & 8 & 8 & 377  \\
\hline
\end{tabular}
\end{table}
At last, we use three hidden layers with the nodes 400-100-50, the accuracy of classification is improved to about 91.4\%. Table \uppercase\expandafter{\romannumeral5} shows the confusion matrix:

\begin{table}[htbp]
\centering
\caption{\label{comparison}The result using three hidden layers with 400-100-50 nodes}
\begin{tabular}{c|c|c|c|c}
\hline
\hline
\multirow{2}{*}{True class 91.4\%} & \multicolumn{4}{|c}{Assigned class}  \\
\cline{2-5}
 & A & L & R & W  \\
\hline
\hline
 A & 770 & 18 & 11 & 1  \\
 \hline
 L & 49 & 342 & 0 & 9  \\
 \hline
 R & 47 & 0 & 345 & 8  \\
 \hline
 W & 5 & 10 & 16 & 369  \\
\hline
\end{tabular}
\end{table}

\subsection{Enhance Classification Accuracy}
\subsubsection{Reject Option}
In order to improve the accuracy of classification, the approach of establishing the "unknown" class is desirable, because there are many fingerprint images with poor quality in the database, such as half-baked ridges by trauma or injury, fingerprint overlap by twice imprint and so on. These fingerprints often lead to the classification error. So we put these fingerprint images with poor quality which can be distinguish by experts into the unknown class. Obviously, with the increasing numbers of fingerprints divided into unknown class, the accuracy of algorithm will significantly increase. But here, we only reject 35 fingerprints (about 1.8\%) which they often result in a large number of errors in the stage of extracting the orientation field and have obvious ambiguity about the types. The accuracy can be increased (Table \uppercase\expandafter{\romannumeral6}):

\begin{table}[htbp]
\centering
\caption{\label{comparison}The result with 1.8\% rejection}
\begin{tabular}{c|c|c|c}
\hline
\hline
 Number of hidden layers (nodes) & 1(600) & 2(400-100) & 3(400-100-50)\\
\hline
\hline
 Accuracy & 92.1\% & 92.6\% & 93.1\%  \\
\hline
\end{tabular}
\end{table}
\subsubsection{Fuzzy Classification}
According to the analysis of section \uppercase\expandafter{\romannumeral4}, we redefine three classes for a fingerprint for convenience, the first one is the real class which the fingerprint belongs to, the second one is the class with the maximum probability value, we call it ¡°the first class¡± and the last one is the class with the second probability value, we call it ¡°the second class¡±. It is likely that the first class is not the real class and the second class is the real one. So a naive idea is that we can set a threshold (e.g. 0.6) for maximum probability value, if it is below the threshold, the fingerprint would be assigned to its second class for next testing. Take the single hidden layer network with 400 nodes for example, the Table \uppercase\expandafter{\romannumeral7} shows the result:
\begin{figure*}[!t]
\centering \subfigure[]{\label{fig:a}\includegraphics[width=0.4\textwidth]{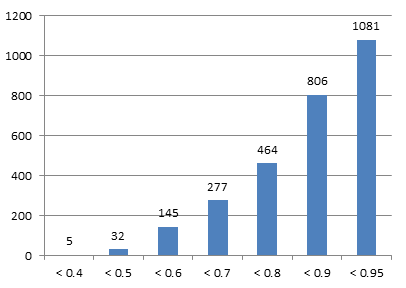}}
\hfill \subfigure[]{\label{fig:b}\includegraphics[width=0.4\textwidth]{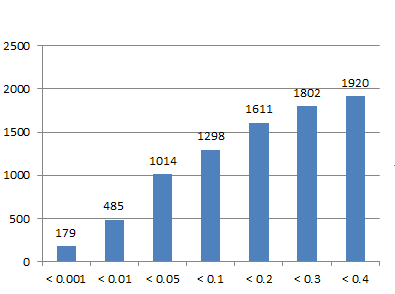}}
\hfill \subfigure[]{\label{fig:c}\includegraphics[width=0.4\textwidth]{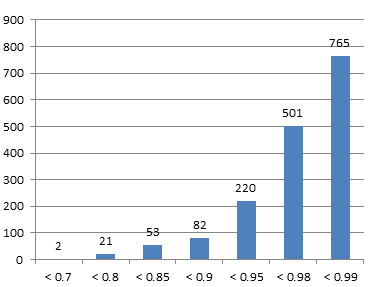}}
\hfill \subfigure[]{\label{fig:c}\includegraphics[width=0.4\textwidth]{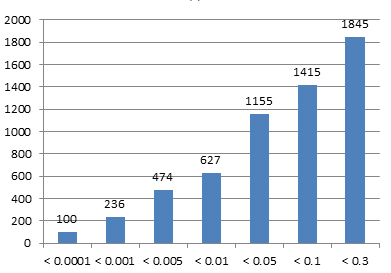}}
\caption{Some statistical results of fuzzy classification. The distribution of (a) the maximum probability value, (b) the second probability value, (c) the sum of two probabilities, (d) the difference of the second and the third probability values. where y-axis := the number of samples, x-axis := the condition of probability value.
} \label{fig:4}
\end{figure*}

\begin{table}[htbp]
\centering
\caption{\label{comparison}The result of fuzzy classification under different threshold value}
\begin{tabular}{c|c|c|c|c|c|c}
\hline
\hline
 threshold & para1 & para2 & para3 & para4 & para5 & acc\\
\hline
\hline
 0.60 & 145 & 83 & 76 & 62 & 2145 & 93.1\%  \\
\hline
 0.70 & 277 & 187 & 127 & 90 & 2277 & 94.3\%  \\
\hline
 0.75 & 347 & 244 & 147 & 103 & 2347 & 94.9\%  \\
\hline
 0.80 & 464 & 346 & 181 & 118 & 2464 & 95.5\%  \\
\hline
 0.85 & 616 & 483 & 214 & 133 & 2616 & 96.1\%  \\
\hline
 0.90 & 806 & 650 & 264 & 156 & 2806 & 97.2\%  \\
\hline
 0.95 & 1801 & 908 & 315 & 173 & 3081 & 98.0\%  \\
\hline
 1.00 & 2000 & 1807 & 360 & 193 & 4000 & 98.8\%  \\
\hline
\end{tabular}
\end{table}

Note: in Table \uppercase\expandafter{\romannumeral7}, para1 is the number of fingerprints with its first probability less than the threshold; para2 is the number of fingerprints with classification success using the maximum probability value under the threshold; para3 is the number of fingerprints with classification success using the second probability value under the threshold; para4 is the number of fingerprints with classification failure using the maximum probability value under the threshold; para5 is the number of fingerprints needing to consider for classification under the threshold; acc is the accuracy of the classification under the threshold.

Through the Table \uppercase\expandafter{\romannumeral7}, the method using simple value is feasible for enhancing the accuracy of classification. Therefore, we can choose one level according to the actual needs for fuzzy classification. In order to further optimize the results, we conduct some statistical analysis of the data (Fig.14 and Table \uppercase\expandafter{\romannumeral8}). Obviously, about half of samples¡¯ maximum probability values are greater than 0.95 and their corresponding categories are usually credible (Fig.14 (a)). We also can obtain the same information in Fig.14 (b). In other words, these samples have stronger ¡°distinguish capacity¡± than other classes. If we only further consider two kinds of probabilities, only 11\% of samples' sum of the first and the second probability values is smaller than 0.95 (Fig.14 (c)). Thus, we should pay special attention to this part of the samples in that the relatively weak ¡°distinguish capacity¡± obtained from classifier. Namely, they may be misclassified twice by fuzzy method using two probabilities. So we can design another condition by considering the sum of two probabilities for ¡°rescuing¡± the error classifying samples (see the next subsection). In conclusion, the classifier is effective for distinguishing fingerprint and fuzzy classification based on the statistical results.

\begin{table}[htbp]
\centering
\caption{\label{comparison}Some results of fuzzy classification}
\begin{tabular}{c|c|c|c|c}
\hline
\hline
 & Min(fp) & Max(sp) & Min(fp + sp) & Max(sp - tp)\\
\hline
\hline
  Value & 0.36 & 0.49 & 0.68 & 0.30 \\
\hline
\end{tabular}
\end{table}

Note: fp := the maximum/first probability value, sp := the second probability value, tp := the third probability value. Min/Max(.) := find the maximum/minimum value.

%\begin{figure*}[htbp]
%\centering \subfigure[]{\includegraphics[width=0.34\textwidth]{16a.png}}
%\hfill \subfigure[]{\includegraphics[width=0.32\textwidth]{16a1.png}}
%\hfill \subfigure[]{\includegraphics[width=0.32\textwidth]{16a2.png}}
%\caption{Samples of misclassification and the mark indicating the region of misclassification. (a) an arch misclassified as a left loop, (b) is the input orientation image, (c) is the reconstructed orientation field by sparse autoencoder.} \label{fig:5}
%\end{figure*}

\begin{figure*}[htbp]
\centering \subfigure[]{\includegraphics[width=0.34\textwidth]{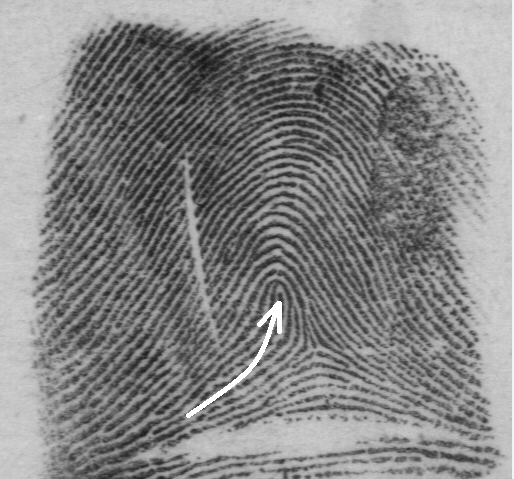}}
\hfill \subfigure[]{\includegraphics[width=0.32\textwidth]{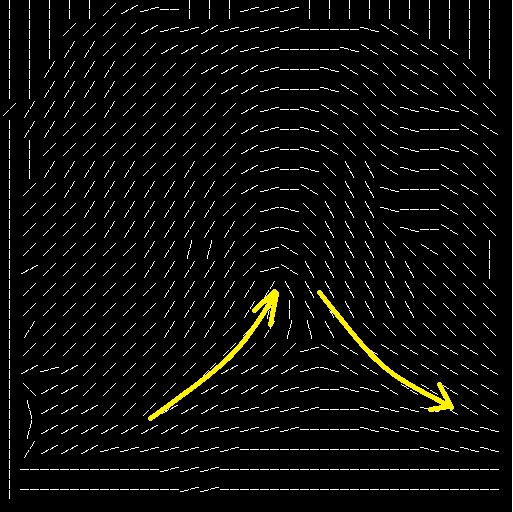}}
\hfill \subfigure[]{\includegraphics[width=0.32\textwidth]{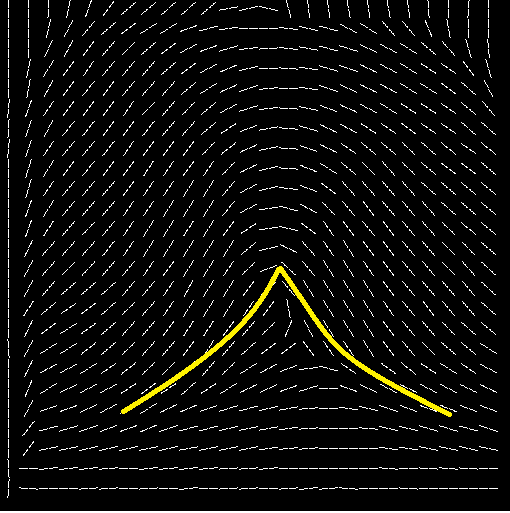}}
\caption{Samples of misclassification and the mark indicating the region of misclassification. (a) a left loop misclassified as an arch, (b) is the input orientation field, (c) is the reconstructed orientation field by sparse autoencoder.} \label{fig:5}
\end{figure*}

\begin{figure*}[htbp]
\centering \subfigure[]{\includegraphics[width=0.34\textwidth]{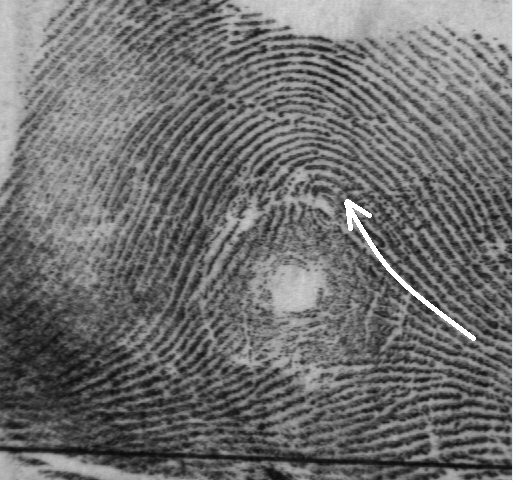}}
\hfill \subfigure[]{\includegraphics[width=0.32\textwidth]{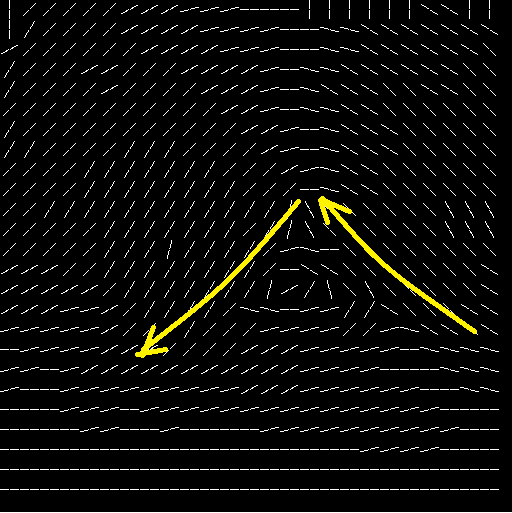}}
\hfill \subfigure[]{\includegraphics[width=0.32\textwidth]{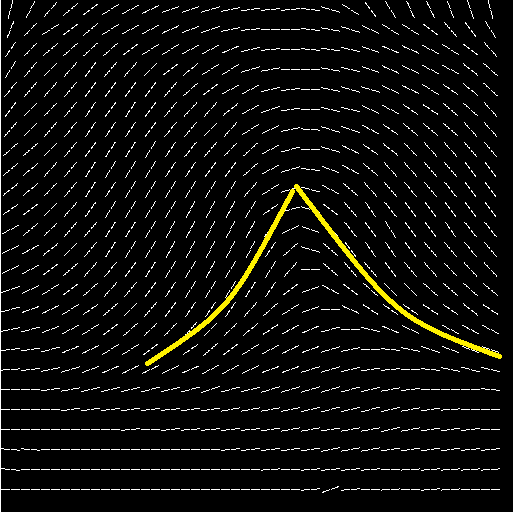}}
\caption{Samples of misclassification and the mark indicating the region of misclassification. (a) a right loop misclassified as an arch, (b) is the input orientation field, (c) is the reconstructed orientation field by sparse autoencoder.} \label{fig:5}
\end{figure*}

\begin{figure*}[htbp]
\centering \subfigure[]{\includegraphics[width=0.34\textwidth]{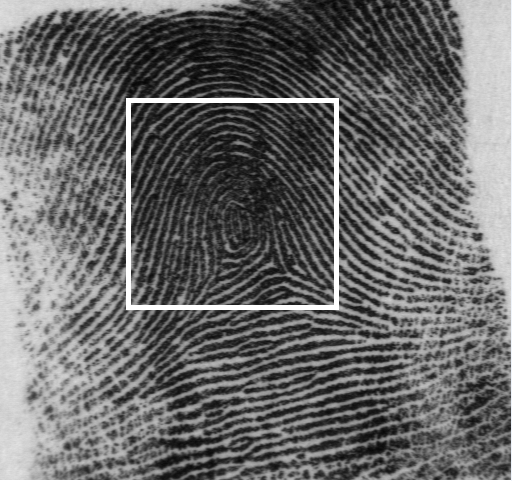}}
\hfill \subfigure[]{\includegraphics[width=0.32\textwidth]{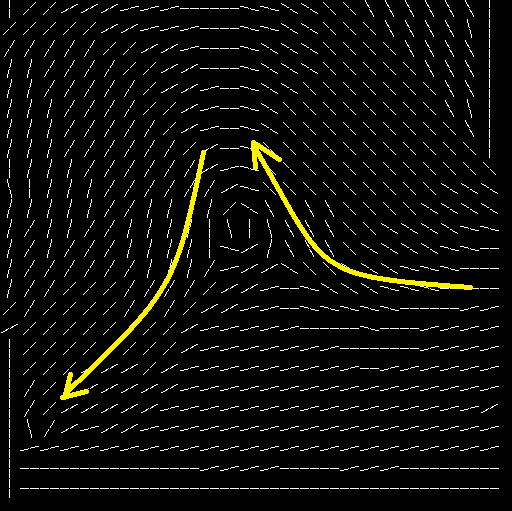}}
\hfill \subfigure[]{\includegraphics[width=0.32\textwidth]{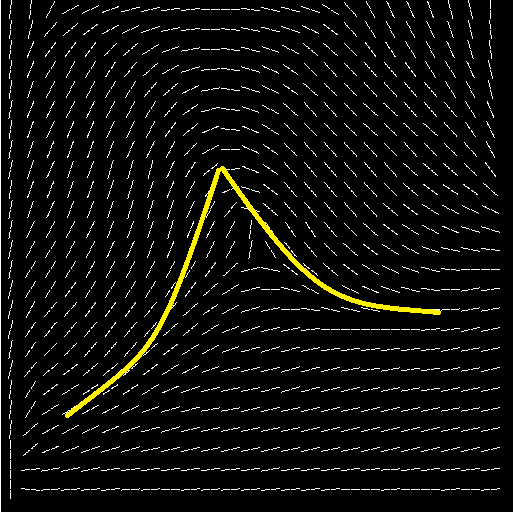}}
\caption{Samples of misclassification and the mark indicating the region of misclassification. (a) a whorl misclassified as an arch, (b) is the input orientation image, (c) is the reconstructed orientation field by sparse autoencoder.} \label{fig:5}
\end{figure*}

\begin{figure*}[htbp]
\centering \subfigure[]{\includegraphics[width=0.34\textwidth]{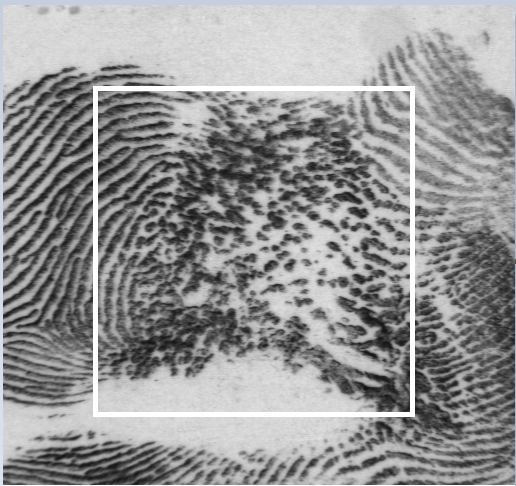}}
\hfill \subfigure[]{\includegraphics[width=0.32\textwidth]{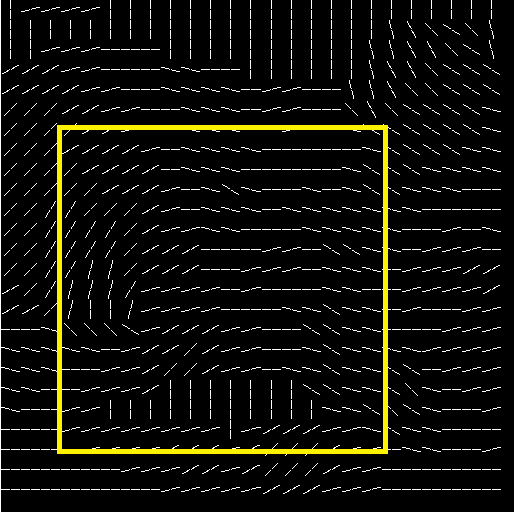}}
\hfill \subfigure[]{\includegraphics[width=0.32\textwidth]{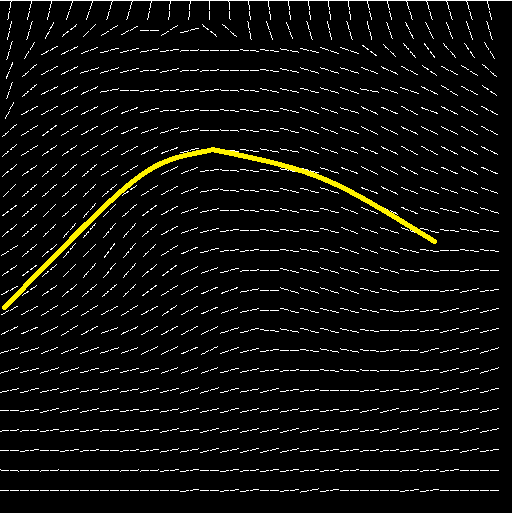}}
\caption{Samples of misclassification and the mark indicating the region of misclassification. (a) a whorl misclassified as an arch, (b) is the input orientation image, (c) is the reconstructed orientation field by sparse autoencoder.} \label{fig:5}
\end{figure*}

\begin{figure}[htbp]
\centering
\subfigure{\includegraphics[width=3.5in]{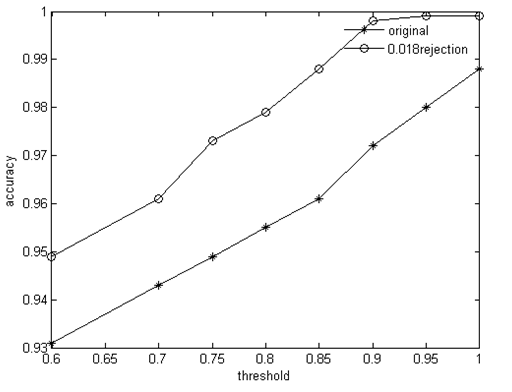}}
\caption{ The result of fuzzy classification with 1.8\% rejection under different thresholds.}
\label{fig_sim}
\end{figure}

\subsection{Misclassification Analysis}
Although we can obtain a better result by using fuzzy method, some samples still can¡¯t be classified into the right type (Fig.15 - Fig.18) when we consider the fuzzy method for the full samples. Through the analysis, we find that the ridge structure or the orientation field we obtained of these samples has a strong ¡°similarity¡± of another type so that the classifier can not recognize the right type or give a high score to their right type. We also observe the probability value of these samples (Table \uppercase\expandafter{\romannumeral8}). If we only use simple condition for fuzzy classification, it is difficult to classify them to their right classes. Of course, the number of these samples is few among the full database.

So, in order to deal with the problem, we try to add a new condition for fuzzy classification. In Table \uppercase\expandafter{\romannumeral9}, we find most of these samples' third probability value indicating their right class. Therefore, we add a new condition: the sum of the first and the second probability values (Fig.14 (c)).

\begin{table}[htbp]
\centering
\caption{\label{comparison}The classification results of samples in Fig.15-17}
\begin{tabular}{c|c|c|c}
\hline
\hline
 & Fig.15 & Fig.16 & Fig.17 \\
\hline
\hline
  A & 0.8771 & 0.8678 & 0.4873 \\
  L & 0.0554 & 0.1182 & 0.2314 \\
  R & 0.0088 & 0.0111 & 0.1586 \\
  W & 0.0587 & 0.0028 & 0.1226 \\
\hline
\end{tabular}
\end{table}

Table \uppercase\expandafter{\romannumeral10} show a simple result about the new condition. Through adding this condition, about 30\% of misclassification samples are recalled.

\begin{table}[htbp]
\centering
\caption{\label{comparison}The result about the new condition for fuzzy classification}
\begin{tabular}{c|c|c|c}
\hline
\hline
  fp + sp & $<0.8$ & $<0.82$ & $<0.85$ \\
\hline
\hline
  num & 21 & 30 & 53 \\
\hline
  recall rate & 29\% & 33\% & 26\% \\
\hline
\end{tabular}
\end{table}

Note: num is the number of samples which are below the threshold.

Also, if we use fuzzy classification for samples with some rejections, we can obtain a better result in ideal cases (Fig.19).

\section{Conclusion}
In this paper, we try to use a novel model, the depth neural network, for fingerprint classification. By the unsupervised feature self-taught learning, we obtain a good reconstruction of the fingerprint which can reflect the types effectively. In order to further improve the accuracy of classification, we adopt the softmax regression for fuzzy classification. According to the statistical analysis of the data, we set different conditions for ameliorating the accuracy of fuzzy classification without rising the costs. We believe that the fuzzy method is very useful in practice. In real life, people always make all possible judgments about uncertain events. Based on this consideration, we provide a secondary class for the "suspicious" fingerprints. The experiments show that our algorithm can get more than 99\% accuracy when we consider the secondary class for each fingerprint. In an era of rapid technological change, the method will not bring huge amount of computation, but greatly improves the accuracy. So, we are convinced that this method is useful and feasible in an actual automatic fingerprint identification system.

In the future, we want to use deep networks to train new feature for classification and try novel types which are different from the traditional five or four classes.

% if have a single appendix:
%\appendix[Proof of the Zonklar Equations]
% or
%\appendix  % for no appendix heading
% do not use \section anymore after \appendix, only \section*
% is possibly needed

% use appendices with more than one appendix
% then use \section to start each appendix
% you must declare a \section before using any
% \subsection or using \label (\appendices by itself
% starts a section numbered zero.)
%

%\appendices
%\section{Proof of the First Zonklar Equation}
%Appendix one text goes here.
%
%% you can choose not to have a title for an appendix
%% if you want by leaving the argument blank
%\section{}
%Appendix two text goes here.
%
%
%% use section* for acknowledgement
\section*{Acknowledgment}
This work was funded by the Chinese National Natural Science Foundation (11331012, 71271204, 11101420).

%The authors would like to thank ...

% Can use something like this to put references on a page
% by themselves when using endfloat and the captionsoff option.
\ifCLASSOPTIONcaptionsoff
  \newpage
\fi

\end{document}